\documentclass[]{spie}  

\usepackage{amsmath,amsfonts,amssymb}
\usepackage{glossaries}
\usepackage{graphicx}
\usepackage{xcolor}
\usepackage{xspace}

\usepackage[colorlinks=true, allcolors=blue, hidelinks]{hyperref}
\usepackage{cleveref}

\newcommand{\etal}[0]{{\em et al.}\xspace}

\newacronym{ae}{AE}{adversarial explanation}
\newacronym{ai}{AI}{artificial intelligence}
\newacronym{ara}{ARA}{accuracy-robustness area}
\newacronym{dst}{DST}{decision support tool}
\newacronym{gmm}{GMM}{Gaussian mixture model}
\newacronym{gpu}{GPU}{graphics processing unit}
\newacronym{idt}{IDT}{intelligent decision tool}
\newacronym{mcts}{MCTS}{Monte-Carlo tree search}
\newacronym{mse}{MSE}{mean-squared error}
\newacronym{nn}{NN}{neural network}
\newacronym{rl}{RL}{reinforcement learning}
\newacronym{rmse}{RMSE}{root-mean-squared error}
\newacronym{ssa}{SSA}{strategically similar autoencoder}
\newacronym{xai}{XAI}{explainable artificial intelligence}
\newacronym{xrl}{XRL}{explainable reinforcement learning}
\newacronym{wm}{WM}{world model}

\title{Combining AI control systems and human decision support via robustness and criticality}

\author[a]{Walt Woods}
\author[b]{Alexander Grushin}
\author[c]{Simon Khan}
\author[d]{Alvaro Velasquez}
\affil[a]{Independent Scholar, USA \texttt{woodswalben@gmail.com}}
\affil[b]{Galois, Inc., USA \texttt{agrushin@galois.com}}
\affil[c]{Air Force Research Laboratory, USA \texttt{simon.khan@us.af.mil}}
\affil[d]{University of Colorado Boulder, USA \texttt{alvaro.velasquez@colorado.edu}}

\usepackage[pscoord]{eso-pic}
\newcommand{\placetextbox}[3]{
\setbox0=\hbox{#3}
\AddToShipoutPictureFG{ \put(\LenToUnit{#1\paperwidth},\LenToUnit{#2\paperheight}){\vtop{{\null}\makebox[0pt][c]{#3}}}
}
}
 
\begin{document} 
\maketitle

\begin{abstract}
AI-enabled capabilities are reaching the requisite level of maturity to be deployed in the real world. Yet, the ability of these systems to always make correct or safe decisions is a constant source of criticism and reluctance to use them. One way of addressing these concerns is to leverage AI control systems alongside and in support of human decisions, relying on the AI control system in safe situations while calling on a human co-decider for critical situations. Additionally, by leveraging an AI control system built specifically to assist in joint human/machine decisions, the opportunity naturally arises to then use human interactions to continuously improve the AI control system’s accuracy and robustness.

We extend a methodology for adversarial explanations (AE) to state-of-the-art reinforcement learning frameworks, including MuZero. Multiple improvements to the base agent architecture are proposed. We demonstrate how this technology has two applications: for intelligent decision tools and to enhance training / learning frameworks. In a decision support context, adversarial explanations help a user make the correct decision by highlighting those contextual factors that would need to change for a different AI-recommended decision. As another benefit of adversarial explanations, we show that the learned AI control system demonstrates robustness against adversarial tampering. Additionally, we supplement AE by introducing strategically similar autoencoders (SSAs) to help users identify and understand all salient factors being considered by the AI system. In a training / learning framework, this technology can improve both the AI's decisions and explanations through human interaction. Finally, to identify when AI decisions would most benefit from human oversight, we tie this combined system to our prior art on statistically verified analyses of the criticality of decisions at any point in time.
\end{abstract}


\section{Introduction}

\placetextbox{.5}{0.090}{\small{Published as:}}
\placetextbox{.5}{0.075}{\small{Walt Woods, Alexander Grushin, Simon Khan, and Alvaro Velasquez}}
\placetextbox{.5}{0.060}{\small{``Combining AI control systems and human decision support via robustness and criticality''}}
\placetextbox{.5}{0.045}{\small{Proc. SPIE 13058, Disruptive Technologies in Information Sciences VIII, 130580J (6 June 2024);}}
\placetextbox{.5}{0.030}{\small{\url{https://doi.org/10.1117/12.3016311}}}

\Gls{rl} can now produce \gls{ai} agents capable of superhuman performance in a wide range of tasks, as evidenced by the recently developed MuZero \cite{schrittwieser2020mastering} and its more sample-efficient successor EfficientZero \cite{ye2021mastering}. Despite these advances, current state-of-the-art \gls{ai} agents can make very costly mistakes in real-world scenarios. For example, an \gls{ai}-controlled car might misinterpret its environment for a few seconds, a mistake that has directly caused loss of life \cite{boudette2021happened}. While these technologies are constantly improving, it is not clear when or if they will ever reach a level of autonomy where they can no longer benefit from collaborating with human decision makers, particularly in environments with decisions on a time scale of hours or days. Effectively, we posit that there will always be critical situations where human/machine teaming will result in more effective decisions than decisions made solely by human or \gls{ai} elements.

A traditional take on human/machine teaming for decision making considers a spectrum of automation, from zero machine involvement to zero human involvement. The automotive industry has codified this idea for autonomous automobiles, where levels are defined as increasing sets of capability that require less human attention \cite{sae2018autonomy_j3016_201806}. 
We instead look at providing a spectrum of {\em support} for human decisions, from minimal decision assistance to maximal decision assistance. This new framing focuses on using \gls{ai} techniques in support of the best overall decisions, regardless of the autonomy capabilities of the \gls{ai} system in isolation; the underlying goal is to maximally empower human decision making, rather than to automate the system as much as possible.

\begin{figure}[t]
    \centering
    \includegraphics[width=\linewidth]{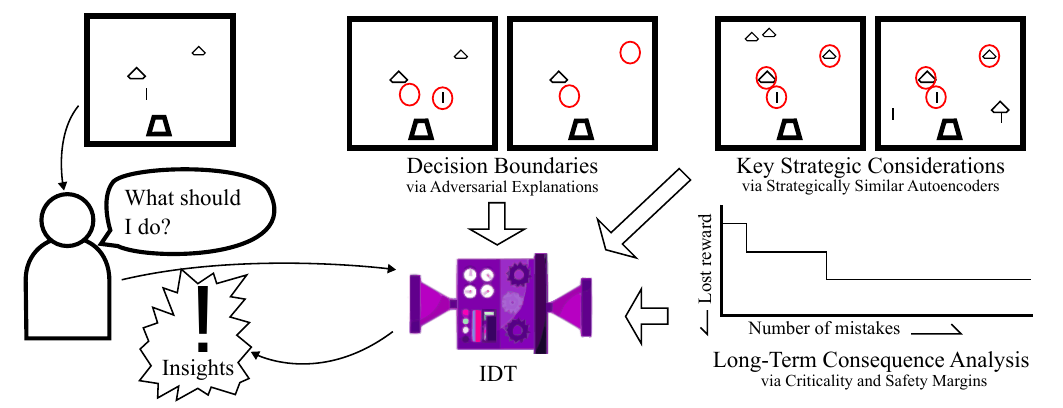}
    \caption{Overview of our \glsentrylong{idt} (\glsentryshort{idt}) effort. By combining base \glsentrylong{rl} (\glsentryshort{rl}) improvements with the developments of \glsentrylongpl{ae} (\glsentryshortpl{ae}), \glsentrylongpl{ssa} (\glsentryshortpl{ssa}), and safety margins, users are shown detailed visual descriptions of key decision components, and provided tools to understand the long-term consequences of potential decisions. Since this information is guided by possible decisions as considered by the AI system, users have the opportunity to explore phenomena that would otherwise be unknown to the user.%
    }
    \label{fig:overview}
\end{figure}

Making good decisions relies, in part, on a good understanding of all available data. This is sometimes called ``evidence-based'' decision making. \Glspl{dst} are an existing class of tool where a user can work on justifying any particular decision by asking specific queries of available data. The answers to those queries can in turn provoke additional questions from the user, and the user can repeat this data interaction loop indefinitely, until they are satisfied that they understand the situation enough to make a well-informed decision. This approach constitutes a relatively minimal level of decision assistance -- it is purely driven by the user, meaning that the user can only explore questions that they know to ask. Colloquially, this means that \glspl{dst} help with known knowns and known unknowns, rather than unknown known/unknown factors. Such an approach has clear limitations when it comes to impactful decision making; notably, there are limitations to the idea of ``evidence-based'' decisions that result from complex environment interactions and unforeseeable events \cite{pawson2011known}. Still, with increasing computing power and availability of high-fidelity simulations, new possibilities arise for providing decision makers with tools that offer much more comprehensive decision assistance.

We propose a new class of tools leveraging \gls{rl} agents that have been trained to make decisions based on available data, and then explaining them through \gls{xai} and related techniques. This approach results in tools that can consider complex environment interactions that might result from decisions made, as opposed to purely data-centric tools like \glspl{dst}. 
The proposed tools can function as a fully autonomous decision system if desired, but importantly, focus on using those decision capabilities to quickly inform human decision-making partners about a situation's full context. This results in a framework that supports both urgent and slow decisions, and can even identify when a human's input in a decision is likely to make a significant difference in mission outcomes.
We call this new class of tools \glspl{idt}.

In this work, we describe methods and techniques for building \glspl{idt} by leveraging prior work \cite{woods2019adversarial,grushin2023safety,grushin2024criticality} and developing new techniques to produce highly detailed and accurate \gls{xai} for state-of-the-art, MuZero-style \gls{rl} agents. An agent training and analysis framework was developed, providing a suite of tools to assist human decision makers in rapidly understanding a scenario and making decisions that fully leverage available data, while also supporting full automation. A brief illustration of the capabilities provided by this system is shown in \cref{fig:overview}, with additional supporting examples in \cref{sec:results}.

\section{Methods}
\label{sec:meth}

To assemble our \gls{idt} prototype, we developed and combined the methods described below. These techniques were designed specifically to produce a general purpose \gls{idt} for helping human decision makers fully leverage the decision-specific knowledge captured by the \gls{rl} agent. To accomplish this, there were four main thrusts to our work:

\begin{enumerate}
    \item General \gls{rl} extensions for building a flexible agent architecture that could be applied to any kind of reinforcement learning problem, while both retaining good learning performance and revealing as much information about the environment as possible.
    \item Extending \gls{ae} \cite{woods2019adversarial} to the \gls{rl} domain to help users better understand different interpretations of available sensor data that would lead to different decision recommendations.
    \item The creation of \glspl{ssa} to help users identify which data is critical to understanding the strategic scenario observed by the \gls{idt}.
    \item Helping users understand the long-term consequences of a decision through criticality (which is defined informally as the reward that an agent can be expected to lose if it makes a certain number of mistakes in a particular situation), and safety margins (the number of mistakes that an agent can afford to make before the expected reward loss is significant)\cite{grushin2023safety,grushin2024criticality}.
\end{enumerate}

Combining all of these methods resulted in an \gls{idt} that could help human users better understand key decision factors and the possible consequences of different decisions in any given situation.

Due to the number of techniques developed as part of this work, many of the techniques will only be briefly described. Much of the presented work is exploratory; further research would be required to fully determine the empirical advantages of our approaches. Relevant citations are provided that will help the reader better understand those parts of the system that are of interest to them.

\subsection{Building a Flexible Agent Architecture}
\label{sec:meth:rl}

MuZero is a high-performing \gls{rl} agent that holds many superhuman performance records \cite{schrittwieser2020mastering}. While promising, that particular agent posed problems for some real-world applications for which, even through simulation, it would be prohibitively expensive to gather the needed 200 million or more data points to train the agent. Fortunately, an agent called EfficientZero presented a few improvements and was also able to surpass human capabilities, using only 100,000 data points \cite{ye2021mastering}. While this is still a significant amount of required data, it made MuZero-style agents feasible for helping to make decisions in many more real-world environments.

At its core, MuZero and EfficientZero both implement explicit \glspl{wm} \cite{ha2018world}. Unlike the original \gls{wm} work, which used game state data at time $t$ to predict the game state at $t+1$, MuZero-style agents use information at time $t$ to predict only the agent's {\em latent representation} of that game state at time $t+1$. This prediction is done implicitly, by optimizing the difference in expected immediate and future rewards from such a rollout. This means that the model can simulate future states, but only through its own strategic, latent interpretation of the world. These latent states are also parameterized by the action selected by the model at each time step, $a_t$. By combining this \gls{wm} simulation capability with \gls{mcts}, a MuZero-style model can explore the consequences of different sequences of actions across multiple steps in time \cite{schrittwieser2020mastering}. While not perfect, the performance of these agents speaks to the method's utility.

While MuZero performs very well, it has a number of limitations. 
The limitations that we thought were most important to address in pursuit of our \gls{idt}, and our explorations into addressing those limitations, are discussed in \cref{sec:app:rl}. Briefly, the resulting improvements include: a new method for balancing pre-LayerNorm activations; a ranked Gaussian method for distributional \gls{rl}; loss segment scaling for less sensitive hyperparameters in different environments; the handling of complex action spaces via a pairwise policy update; direct control of the exploration/exploitation trade-off via a pairwise policy update; a particle swarm tree search alternative to \gls{mcts}; and a modified MuZero value representation for easier model learning.

With these enhancements in hand, the \gls{idt}'s agents successfully learned to execute tasks in a variety of environments.

\subsection{Finding Relevant Decision Boundaries with Adversarial Explanations}

\Glspl{ae} are a technique for exploring a large number of counterfactuals to help better understand any given decision from an \gls{ai} system. This is accomplished by conditioning \glspl{nn} to be queried for decision boundaries that are close to a given input. For a full description of \gls{ae}, see the work which established the technique \cite{woods2019adversarial}, and applied it to neural networks that were trained in a supervised way.

In this work, we extended \gls{ae} to \gls{rl}. As discussed in \cref{sec:meth:rl}, this was done in the context of a MuZero-style agent, meaning that we also wanted explanations to work in the context of the dynamics function. Adapting the conditioning step of \gls{ae} worked as follows: as in prior work \cite{woods2019adversarial}, one parameter was chosen for each example in the training batch, using a random rollout time $\tau$, and a random choice from the available $Q$ and policy network outputs. The gradient of this parameter with respect to the inputs was computed, and some function of that gradient was added to the overall loss function, as in previous approaches \cite{woods2019adversarial}.

Multiple additions to the basic approach helped  optimize performance in an \gls{rl} setting. These were gradient scale correction, optimizer scale correction, a different approach for controlling the strength of the regularization, and a sparsity-promoting gradient-minimization function. Details of these methods may be found in \cref{sec:app:ae}.

\begin{figure}
    \centering
    \includegraphics[width=\linewidth]{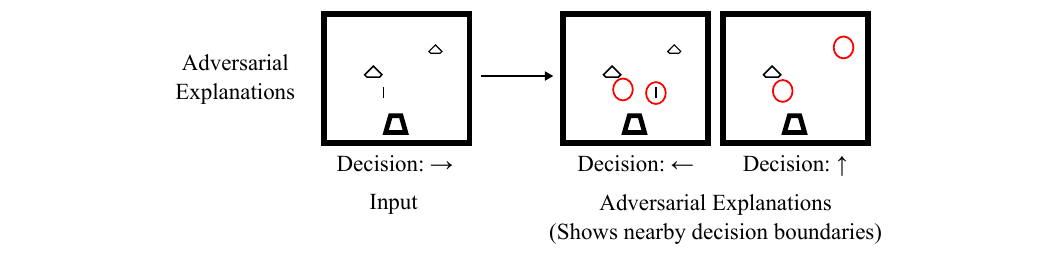}
    \caption{\Glsentrylongpl{ae} (\glsentryshortpl{ae}) provide a methodology for hardening \gls{ai} systems against subtle manipulations and then using the hardened systems to find accurate, nearby decision boundaries that are semantically meaningful to human observers.  The leftmost frame represents the original, unperturbed input; the agent chooses to move to the right, in order to avoid the obstacle to its left. The middle frame shows a perturbation to the input, made such that the agent chooses to move to the left instead; here, the obstacle has shifted to the right, as illustrated via red circles.  The rightmost frame shows a perturbation made in order for the agent to keep moving forward (rather than left or right); here, the obstacle disappears altogether.  These perturbations suggest that the agent's decision is influenced directly by the obstacle.}
    \label{fig:meth:ae:overview}
\end{figure}

Together, these improvements successfully allowed \gls{ae} to be applied to \gls{rl}. Highlights of \gls{ae} in an \gls{rl} context are shown in \cref{fig:meth:ae:overview}.

\subsection{Identifying Critical Data with Strategically Similar Autoencoders}

\begin{figure}[t]
    \centering
    \includegraphics[width=\linewidth]{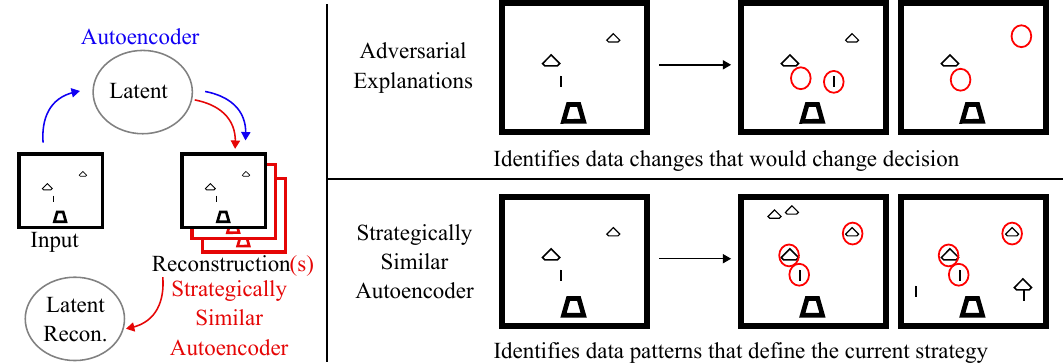}
    \caption{\Glsentrylongpl{ssa} (\glsentryshortpl{ssa}) differ from standard autoencoders -- rather than reconstructing the input, they reconstruct a family of inputs which the AI interprets as the same (reconstructing the same latent code). Through this process, data patterns that define the current strategy can be identified. These patterns are complementary to the patterns identified through \glspl{ae}.  As for \glspl{ae}, the leftmost frame represents the original input, while the middle and right frames represent two different reconstructions of the input.  Note that the obstacle and two enemy ships persist in both reconstructions, suggesting that they are important to the agent.}
    \label{fig:meth:ssa:overview}
\end{figure}

In the original paper on \glspl{wm}, Ha and Schmidhuber used autoencoders to compress the agent's sensors into a latent representation that could be re-expanded to the original sensor representation \cite{ha2018world}. This classic use of autoencoders is a great way for identifying the most visually significant varying factors in sensory space, and compressing them. However, qualities of the input are retained based on visual similarity instead of their use in the task being handled by the \gls{ai} system.
We note that MuZero's adaptation of the \gls{wm} work stopped propagating this explicitly reconstructable representation, instead focusing on propagating enough state to predict action probabilities and corresponding reward values \cite{schrittwieser2020mastering}. From an agent efficacy point of view, this is advantageous, as exactly reconstructing the sensor inputs is irrelevant to task performance. 

For an \gls{idt} whose goal is to help human/machine teams make good decisions, reconstructing some version of the sensor input might be very useful. Specifically, reconstructing the parts of the input that the \gls{ai} considers strategically important to any potential decisions would allow for a human to better understand which environmental factors the \gls{ai} considers salient, and which ones it ignores. To address this, we developed \glspl{ssa}.

The core insight for \glspl{ssa} is that an autoencoder does not necessarily need to reconstruct the input. Instead, the same architecture may be used to reconstruct the latent decision space that MuZero uses to make decisions. That is, rather than using the classic autoencoder setup of inputs $\to$ latent $\to$ inputs, we instead follow latent $\to$ inputs $\to$ latent. This problem is fundamentally underconstrained; that is, by design, multiple inputs often map to a single latent code, as the differences between those inputs are not relevant to the \gls{rl} agent's task. Importantly, this function can be completely detached from the main \gls{rl} agent's learning process, giving us insight into the model without the need to affect the model's learning. 

To implement this, we combined ideas from StarGAN \cite{choi2020stargan}, WassersteinGAN \cite{arjovsky2017wasserstein}, and VEEGAN \cite{srivastava2017veegan}. Briefly, the standard generator/discriminator relationship is trained as normal. However, using VEEGAN, we concatenate additional values to the latent code specifically for capturing variances not represented by the latent code. These variances are trained to not be useful in real/fake determination by making real examples look fake (similar to VEEGAN \cite{srivastava2017veegan}, but we found it easier to optimize this objective when using predicted variance for both real and fake examples). To ensure that these variances capture significant visual variety within the input, the style diversification loss from StarGAN is included \cite{choi2020stargan}. With the latent code being included as input to VEEGAN, we noticed that VEEGAN could have a difficult time learning useful variances with the variance vector, mostly relying on the latent code. While the style diversification loss helps with this, we also introduced a version of L2 loss that supports Brownian motion of L2-style predictions (by default, L2 predicts the mean, which for the variance vector would be all zeros). We used a simple bias term to translate between L1- and L2-style loss, as $\frac{\beta (a - b)^2}{\beta^2 + d(a - b)^2}$, with $d$ being the detachment operator. Finally, this loss was divided at each point by the likelihood of the sampled fake variance, allowing for the definition of each variance number to drift when those variances cannot effectively be predicted, but also allowing the network to learn to reconstruct specific variance numbers once they can be predicted.

The result is that the strategic scenario -- and variations of that scenario -- can be reconstructed for the user within the \gls{idt}. These reconstructions allow the user to explicitly see which environmental factors are recognized as potentially strategically significant by the \gls{rl} agent. This is illustrated in \cref{fig:meth:ssa:overview}.

\subsection{Flagging Long-Term Consequences with Criticality and Safety Margins}
\label{sec:meth:crit}

From an \gls{idt} point of view, \gls{wm}-style simulation provides the ability to explore the potential long-term effects of different decisions. The \Gls{ssa} allows these simulated counterfactuals to be cast back into a sensory space, allowing for easier understanding of which strategic elements are being tracked. \Glspl{ae} allow for such sensory space interpretations to be modified, to better understand how minor initial differences in the scenario might lead to wildly different long-term outcomes. However, while these tools allow for the investigation of long-term consequences through the agent's \gls{wm}, such an investigation is often expensive in terms of the human decision partner's time. 

To help focus human decision maker time on only those decisions that are truly critical, we propose using criticality and safety margins, which we derive using recently published techniques \cite{grushin2023safety,grushin2024criticality}. These approaches use statistics to derive bounds for any decision the agent must make. Essentially, safety margin bounds say that an agent might make up to $N_{safety}$ random (potentially erroneous) actions before there is a reasonable chance that task performance might be negatively affected beyond some user-defined threshold. 

\begin{figure}[t]
    \centering
    \includegraphics[width=\linewidth]{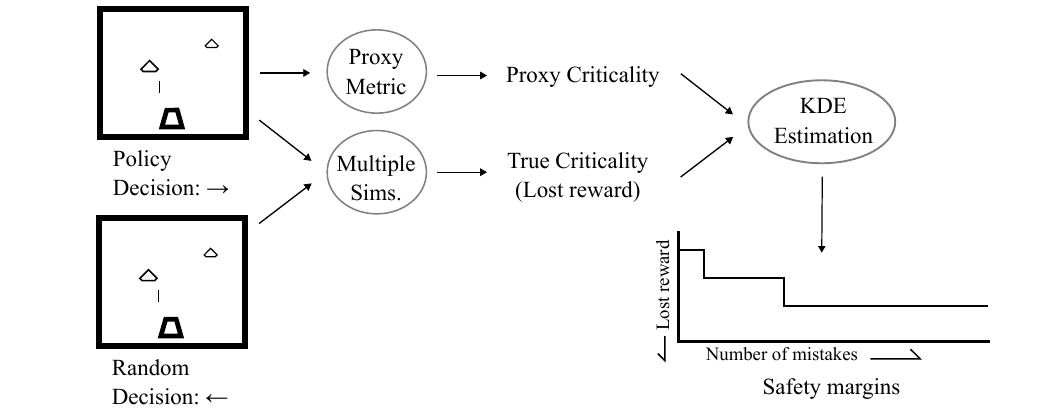}
    \caption{Safety margins (shown in the bottom right corner of the figure) capture the relationship between the number of consecutive mistakes (more formally, random actions) that an agent might hypothetically make and the resulting expected loss of reward (note that higher reward losses appear lower along the vertical axis).  To derive this relationship at some time $t$, we first run a simulation beginning at $t$, and measure the total discounted reward collected.  We then repeat the simulation multiple times, but for some number of time steps $n$ beginning with time step $t$, replace the actions output by the agent's policy with randomly-chosen actions, and measure the mean loss (reduction) in reward as a result of these random actions; this is the true criticality at time $t$.  We also use a much faster, but typically much less accurate approach called a proxy criticality metric to compute the proxy (approximate) criticality at time $t$ without running any simulations  \cite{lin2017}.  This aforementioned process is repeated offline for many episodes, at different time steps, in order to collect a dataset containing proxy and true criticality values.  From this dataset, kernel density estimation is performed to capture the statistical relationship between proxy and true criticality for each value $n$.  Finally, the kernel density estimates are used to compute a table of safety margins.  When the agent is deployed, only proxy criticality needs to be computed at any given time $t$; it is then used to efficiently look up the safety margins for $t$ within the table, without the need to run additional simulations.}
    \label{fig:meth:crit:overview}
\end{figure}

Prior work shows that, in the Atari game Beamrider for example, $47\%$ of agent losses could potentially be prevented by looking at only the lowest $5\%$ of safety margins \cite{grushin2023safety,grushin2024criticality}. In other words, when a human decision maker's time is limited, safety margins allow them to spend more time looking at a smaller number of critical decisions. An illustration of safety margins and criticality is shown in \cref{fig:meth:crit:overview}.

\section{Results}
\label{sec:results}

We focus on qualitative results, walking through a limited number of examples from the Atari game Beamrider. Additional non-public examples exist that apply these ideas to, e.g., scenarios in the AFSIM simulation framework. We encourage prospective partners who are interested in using \glspl{idt} for human/machine decision teaming to contact us.

\subsection{Model Criticality}
Criticality requires one measurement that is dependent on the model: proxy criticality \cite{grushin2023safety,grushin2024criticality}. For our modified MuZero-style agents, we use $E[\text{reward} | \text{policy}] - E[\text{worst reward} | \text{1 wrong step}]$, i.e., the difference between the anticipated reward if the best action (i.e., the action prescribed by the policy) is taken and the anticipated reward if the worst action is taken \cite{lin2017}. While there are likely better proxy criticality formulas \cite{grushin2024criticality}, especially when considering safety margins of greater than one step, this was sufficient for initial results.

\begin{figure}[t]
    \centering
    \includegraphics[height=3.5in]{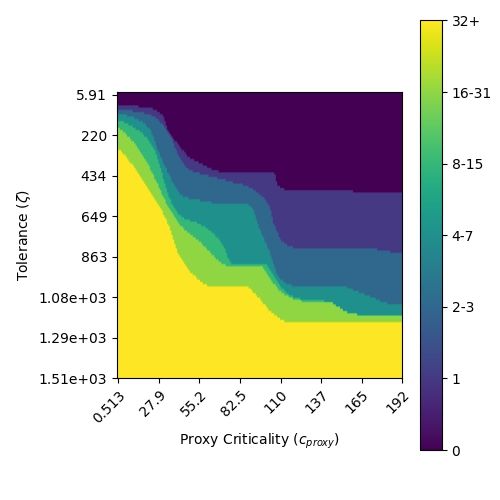}
    \caption{Safety margins (color) given a specified acceptable loss in mission outcome (y-axis) and proxy criticality (x-axis).}
    \label{fig:results:model:sm}
\end{figure}

\begin{figure}[t]
    \centering
    \includegraphics[height=1.2in]{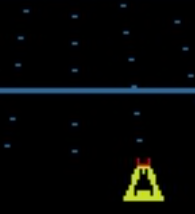}
    \includegraphics[height=1.2in]{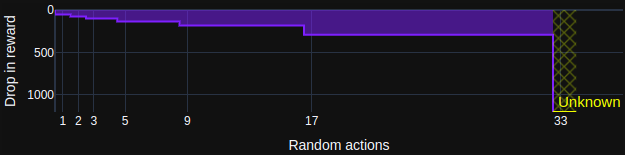}
    \\ (A) \\
    \includegraphics[height=1.2in]{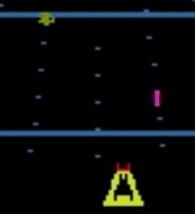}
    \includegraphics[height=1.2in]{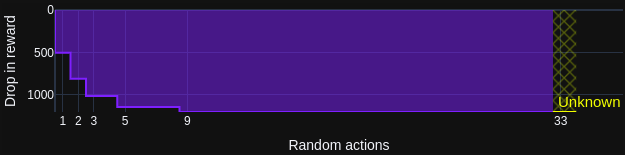}
    \\ (B) \\
    \caption{To present the safety margins from \cref{fig:results:model:sm} to the user, the proxy criticality at the current time step $t$ is looked up on the horizontal axis, and a vertical slice is shown. (A) shows a situation with proxy criticality 2.15, and (B) shows a situation with proxy criticality 152.65. }
    \label{fig:results:model:sm-applied}
\end{figure}

For an agent trained on Beamrider, the resulting safety margin plot is shown in \cref{fig:results:model:sm}, with supporting kernel density estimates from the collected data shown in \cref{fig:app:model:kde}. For an example of how the safety margin plot is applied and shown to the user in various scenarios, see \cref{fig:results:model:sm-applied}.

\subsection{Scenario 1: Obstacle Approaching from Left}

\begin{figure}[ht!]
    \centering
    \includegraphics[height=0.19\linewidth]{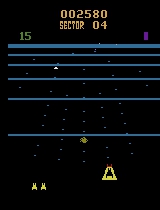}
    \begin{minipage}{0.1\linewidth}
        \centering
        \vspace*{-1.85\linewidth}
        $\to$
    \end{minipage}
    \includegraphics[height=0.19\linewidth]{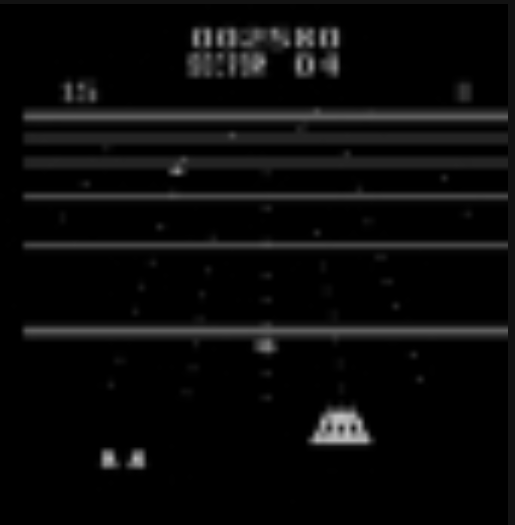} \\
    Visual input to a human $\to$ Input to \gls{ai} \\
    (23.4 proxy criticality) Left 3\% vs Right 97\%.
    Safety margin 4 at 220, 16 at 440 \\
    
    \includegraphics[width=0.59\linewidth]{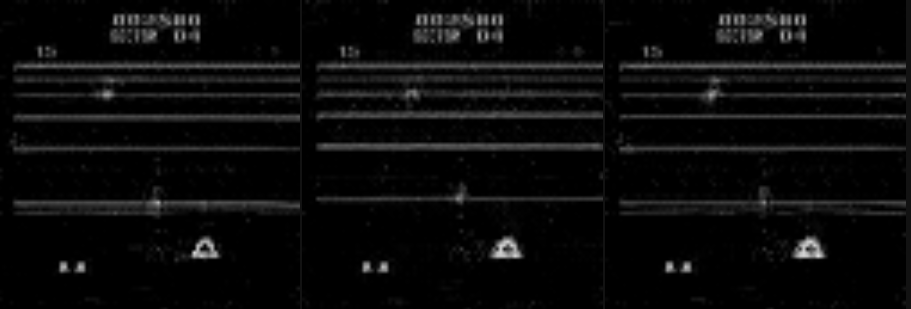} \\
    \Gls{ssa} frames 
    \vspace*{0.2em}\\

    \begin{tabular}{cc}
        \includegraphics[width=0.19\linewidth]{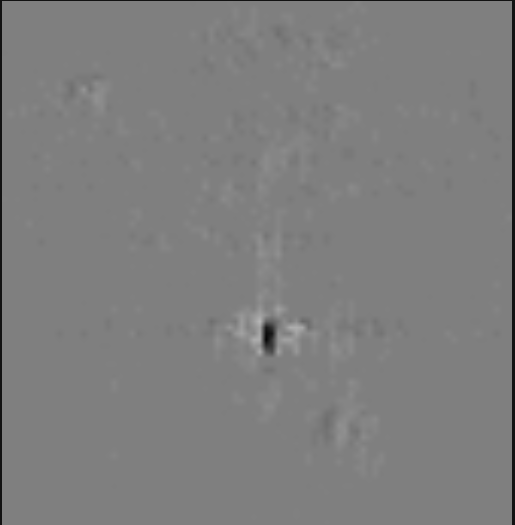}
        \includegraphics[width=0.19\linewidth]{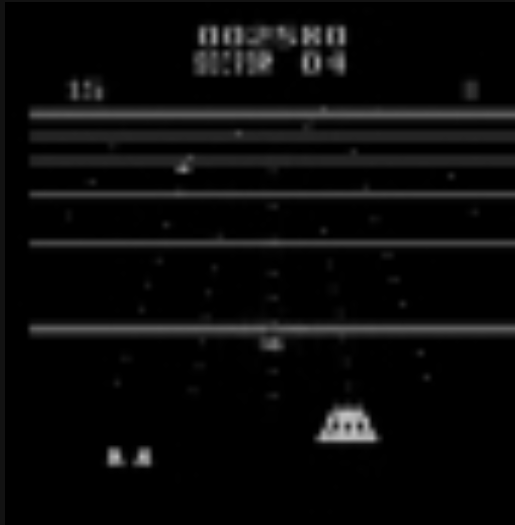}
        & \includegraphics[width=0.19\linewidth]{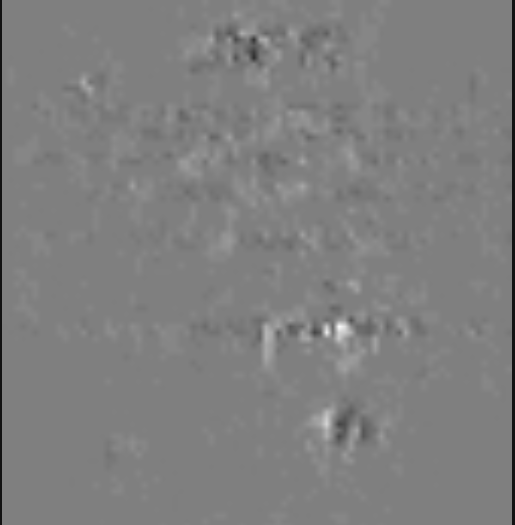}
        \includegraphics[width=0.19\linewidth]{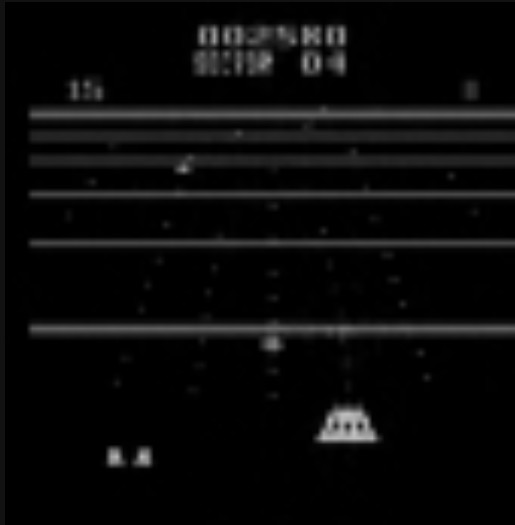}
        \\ \multicolumn{2}{c}{Adversarial explanations for ``Left'' and ``Right'' actions, original \gls{ae} evaluation.}
        
        \\ \includegraphics[width=0.19\linewidth]{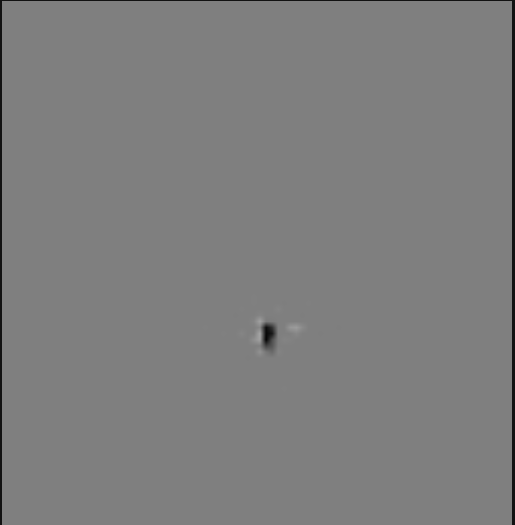}
        \includegraphics[width=0.19\linewidth]{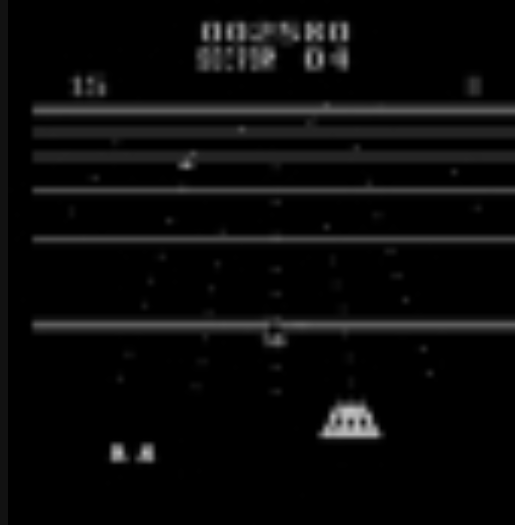}
        & \includegraphics[width=0.19\linewidth]{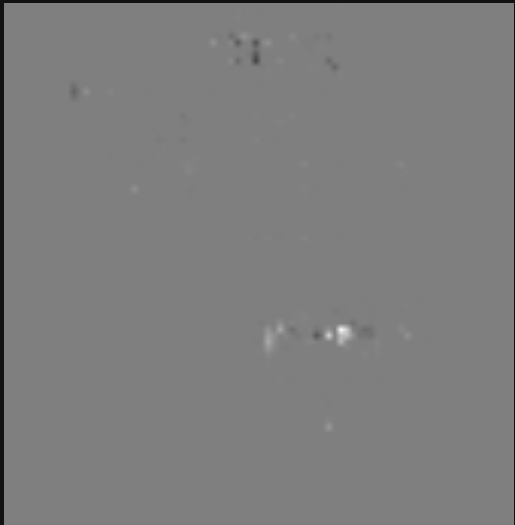}
        \includegraphics[width=0.19\linewidth]{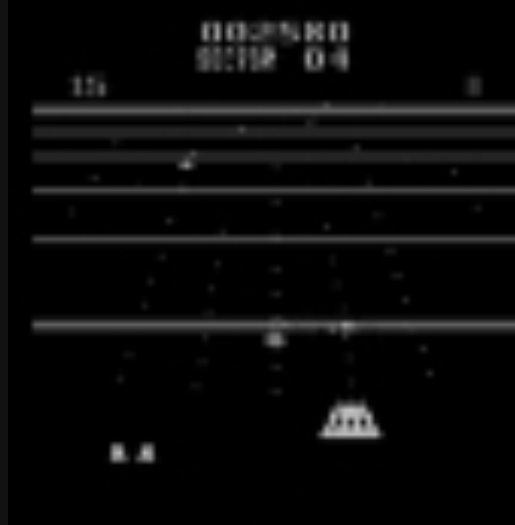}
        \\ \multicolumn{2}{c}{Adversarial explanations for ``Left'' and ``Right'' actions, sparsity-promoting evaluation (\cref{sec:app:ae:sparsity}).}
    \end{tabular}

    \vspace*{0.5em}
    \caption{Case study of analyzing decisions with an obstacle approaching from the left. The SSA shows a surprising reliance of this particular model on the specific gridline positions (note that these positions are similar across the three reconstructions) -- this could be due to the proximity of the observed obstacle and one of the gridlines; see \cref{fig:results:s3} and \cref{fig:results:s4} for examples where the gridline locations were not strategically significant. This reliance on gridline positions can be viewed as an unknown unknown factor \cite{pawson2011known}.  For adversarial explanations, on the left side of the figure, we show an adversarial perturbation that encourages the model to decide to move ``Left''; the leftmost frame is the difference between the original perturbed input (the second frame from the left) and the unperturbed input (shown at the top right of the figure, above ``Input to AI'').  The model correctly identifies the removal of the obstacle (note the dark patch in the leftmost image) as justification for going ``Left''.  Similarly, on the right side of the figure, the perturbation creates additional obstacles (shown as light patches in the difference between the perturbed and unperturbed input) in front of the player to further encourage a decision to move ``Right''.}
    \label{fig:results:s1}
\end{figure}

This scenario is demonstrated in \cref{fig:results:s1}.

\subsection{Scenario 2: Obstacle Approaching from Right}

\begin{figure}[ht!]
    \centering
    \includegraphics[height=0.19\linewidth]{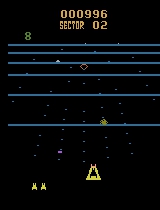}
    \begin{minipage}{0.1\linewidth}
        \centering
        \vspace*{-1.85\linewidth}
        $\to$
    \end{minipage}
    \includegraphics[height=0.19\linewidth]{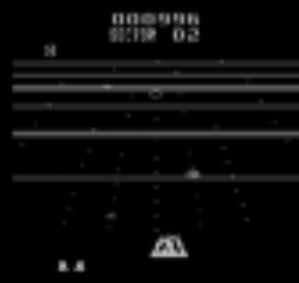} \\
    Visual input to a human $\to$ Input to \gls{ai} \\
    (28.7 proxy criticality) Left 99.9\% vs Right 0.1\%.
    Safety margin 4 at 326, 16 at 504 \\
    
    \includegraphics[width=0.59\linewidth]{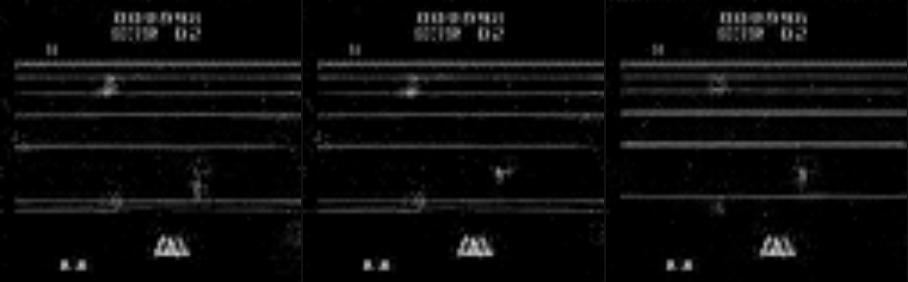} \\
    \Gls{ssa} frames 
    \vspace*{0.2em}\\

    \begin{tabular}{cc}
        \includegraphics[width=0.19\linewidth]{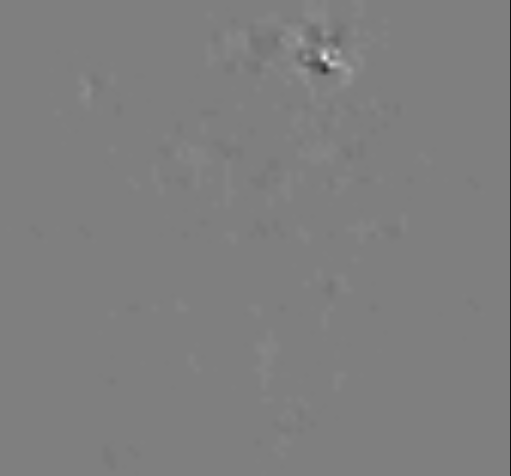}
        \includegraphics[width=0.19\linewidth]{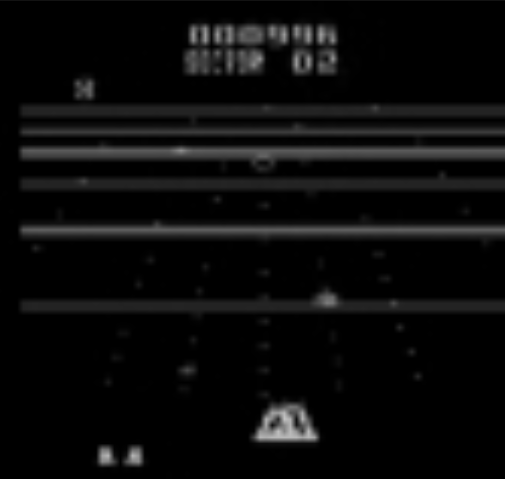}
        & \includegraphics[width=0.19\linewidth]{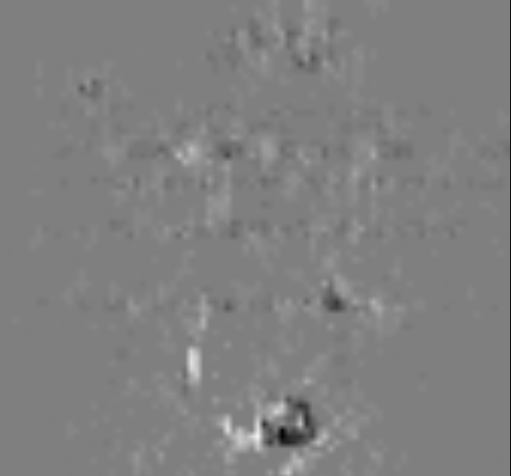}
        \includegraphics[width=0.19\linewidth]{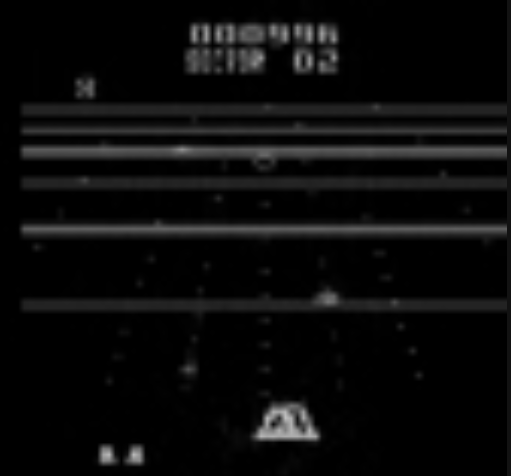}
        \\ \multicolumn{2}{c}{Adversarial explanations for ``Left'' and ``Right'' actions, original \gls{ae} evaluation.}
        
        \\ \includegraphics[width=0.19\linewidth]{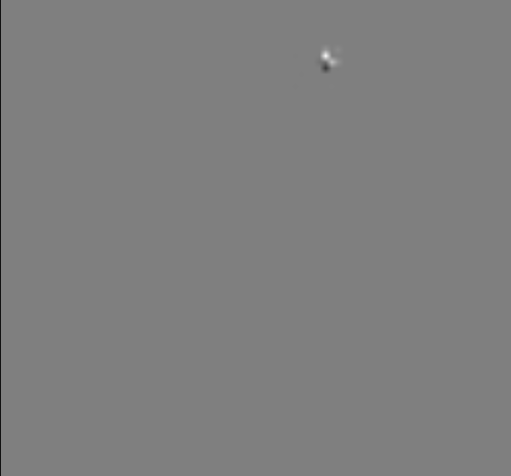}
        \includegraphics[width=0.19\linewidth]{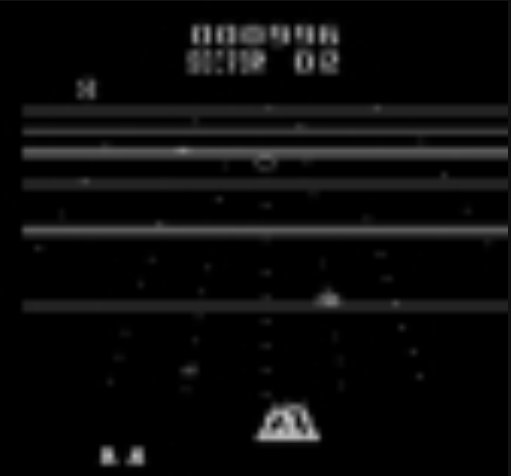}
        & \includegraphics[width=0.19\linewidth]{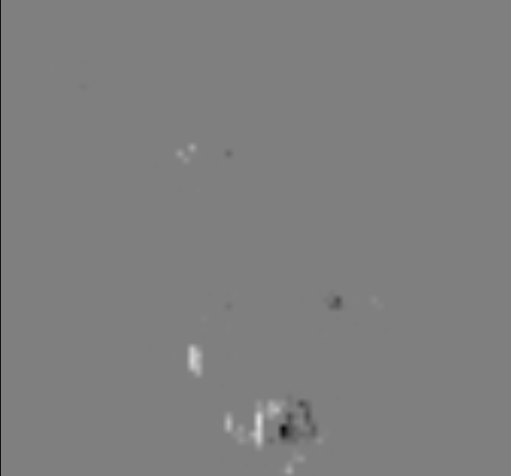}
        \includegraphics[width=0.19\linewidth]{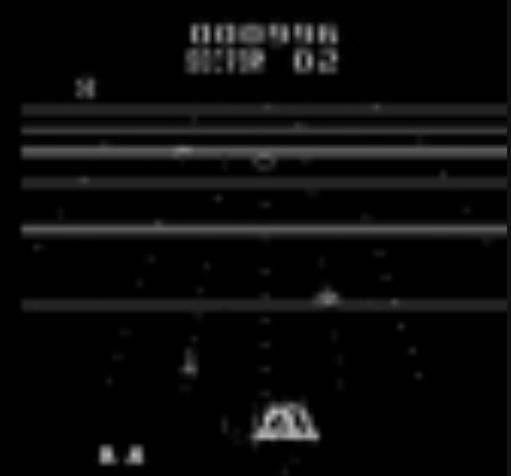}
        \\ \multicolumn{2}{c}{Adversarial explanations for ``Left'' and ``Right'' actions, sparsity-promoting evaluation (\cref{sec:app:ae:sparsity}).}
    \end{tabular}

    \vspace*{0.5em}
    \caption{Case study of analyzing decisions with an obstacle approaching from the right (with the obstacle on the left no longer posing a threat). Similar to \cref{fig:results:s1}, the \gls{ssa} reconstructions indicate a reliance on gridline positions. Surprisingly, \gls{ae} alters the ``Sector'' number of the game information to further encourage a ``Left'' action; this can be viewed as another unknown unknown factor that was uncovered. We note that this might be a result of the already saturated action distribution: the agent already favors a ``Left'' action, but the action might be even more favored in a different sector. The sparsity-promoting evaluation method shows that a ``Right'' action would make sense if the object to the right was not dangerous and there was an obstacle on the left.}
    \label{fig:results:s2}
\end{figure}

This scenario is demonstrated in \cref{fig:results:s2}.

\subsection{Scenario 3: No Immediate Action Required}

\begin{figure}[ht!]
    \centering
    \includegraphics[height=0.19\linewidth]{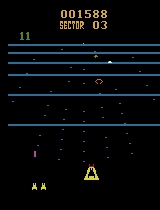}
    \begin{minipage}{0.1\linewidth}
        \centering
        \vspace*{-1.85\linewidth}
        $\to$
    \end{minipage}
    \includegraphics[height=0.19\linewidth]{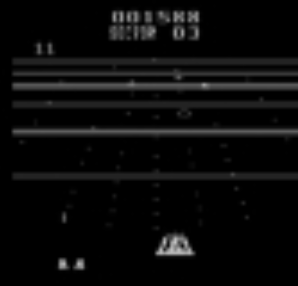} \\
    Visual input to a human $\to$ Input to \gls{ai} \\
    (2.95 proxy criticality) Left 52\% vs Right 48\%.
    Safety margin 4 at 101, 16 at 184 \\
    
    \includegraphics[width=0.59\linewidth]{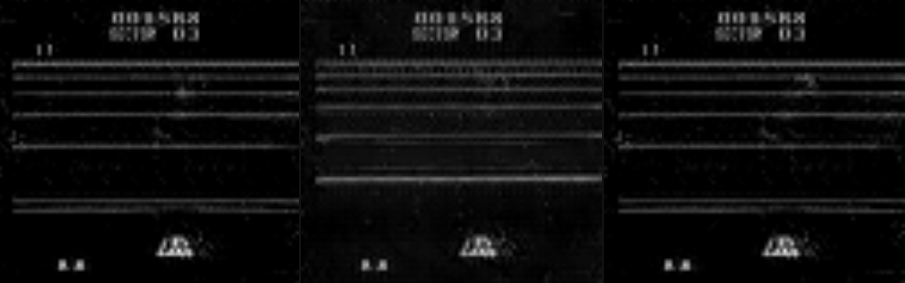} \\
    \Gls{ssa} frames 
    \vspace*{0.2em}\\

    \begin{tabular}{cc}
        \includegraphics[width=0.19\linewidth]{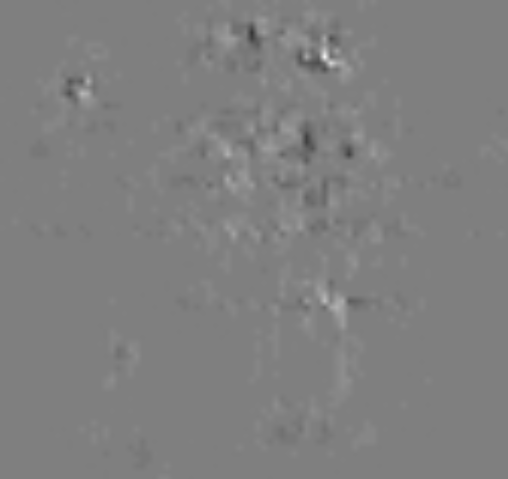}
        \includegraphics[width=0.19\linewidth]{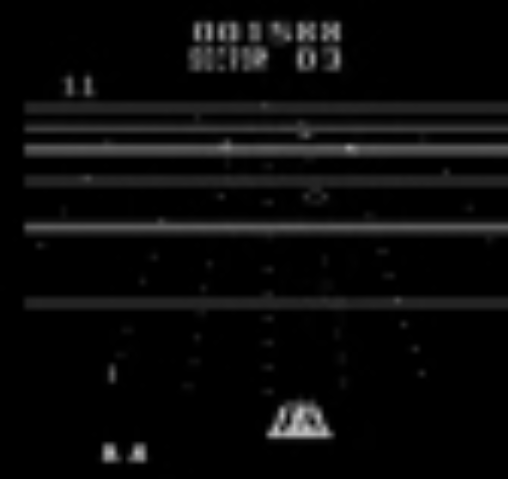}
        & \includegraphics[width=0.19\linewidth]{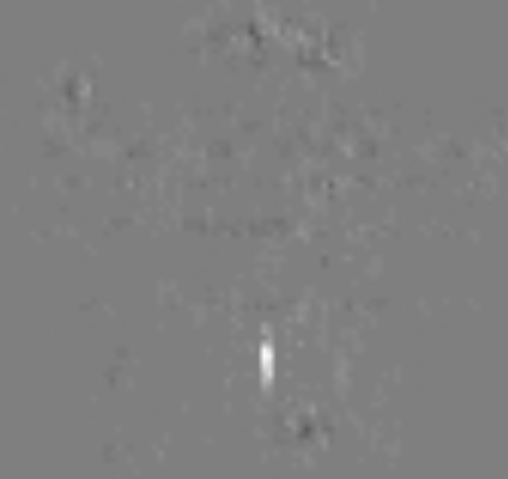}
        \includegraphics[width=0.19\linewidth]{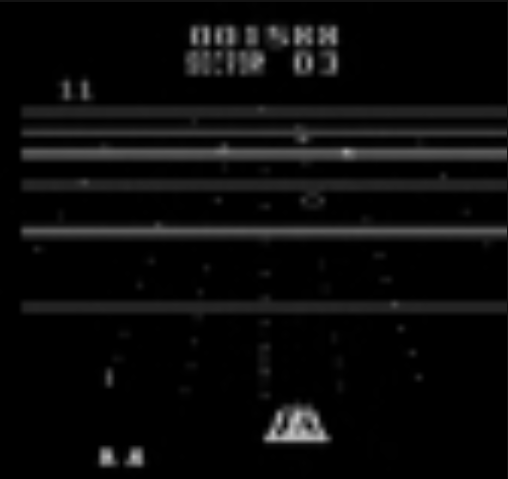}
        \\ \multicolumn{2}{c}{Adversarial explanations for ``Left'' and ``Right'' actions, original \gls{ae} evaluation.}
        
        \\ \includegraphics[width=0.19\linewidth]{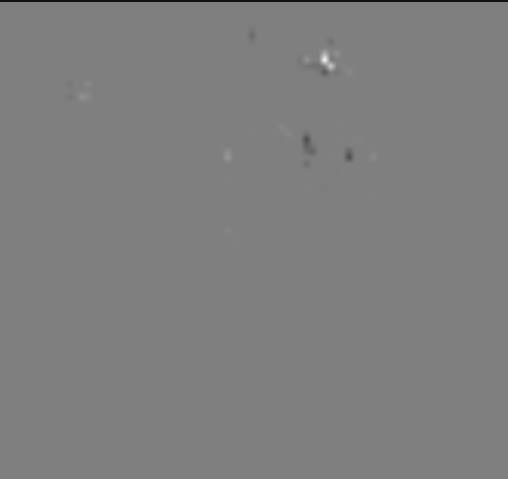}
        \includegraphics[width=0.19\linewidth]{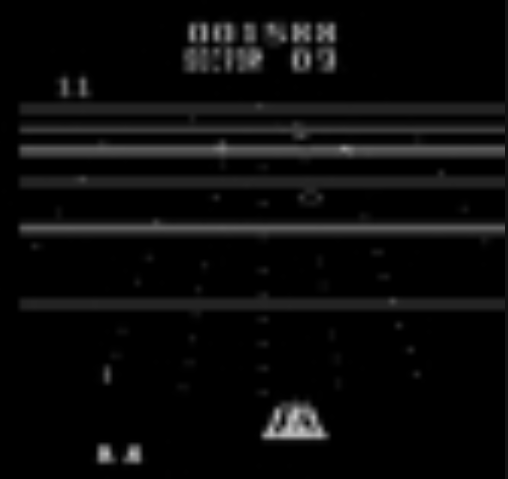}
        & \includegraphics[width=0.19\linewidth]{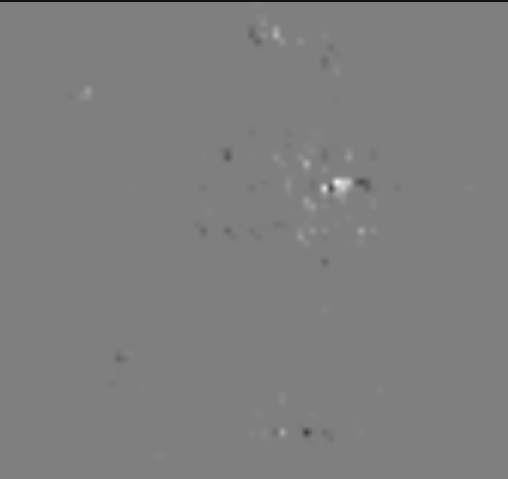}
        \includegraphics[width=0.19\linewidth]{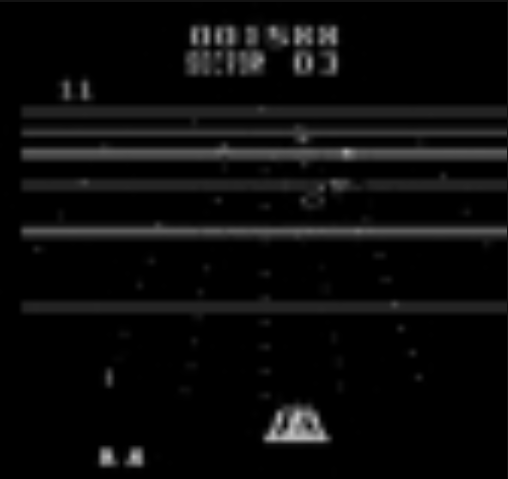}
        \\ \multicolumn{2}{c}{Adversarial explanations for ``Left'' and ``Right'' actions, sparsity-promoting evaluation (\cref{sec:app:ae:sparsity}).}
    \end{tabular}

    \vspace*{0.5em}
    \caption{Case study of analyzing decisions with no immediate concerns. Here, the \gls{ssa} reconstructions show non-reliance on the gridlines, with significant variations in their positions across the three reconstructions. \Glspl{ae} for ``Left'' and ``Right'' show a reliance on sector information (for the ``Left'' action in particular) and different potential targets and obstacles being observed.}
    \label{fig:results:s3}
\end{figure}

This scenario is demonstrated in \cref{fig:results:s3}.

\subsection{Scenario 4: Non-Gridline-Aligned Obstacle}

\begin{figure}[ht!]
    \centering
    \includegraphics[height=0.19\linewidth]{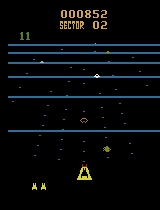}
    \begin{minipage}{0.1\linewidth}
        \centering
        \vspace*{-1.85\linewidth}
        $\to$
    \end{minipage}
    \includegraphics[height=0.19\linewidth]{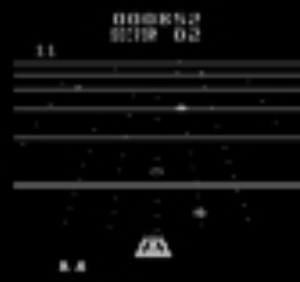} \\
    Visual input to a human $\to$ Input to \gls{ai} \\
    (20.87 proxy criticality) Left 98\% vs Right 2\%.
    Safety margin 4 at 172, 16 at 374 \\
    
    \includegraphics[width=0.59\linewidth]{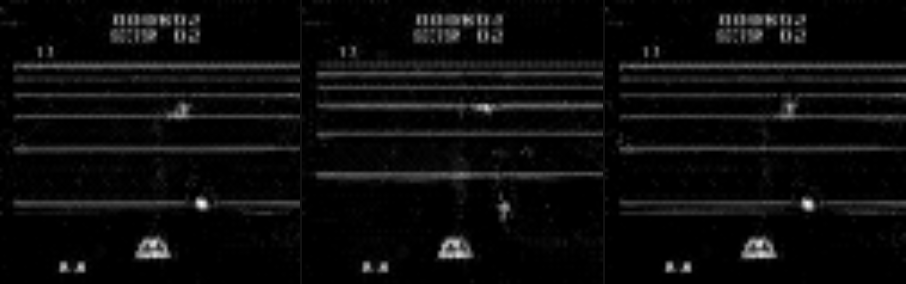} \\
    \Gls{ssa} frames 
    \vspace*{0.2em}\\

    \begin{tabular}{cc}
        \includegraphics[width=0.19\linewidth]{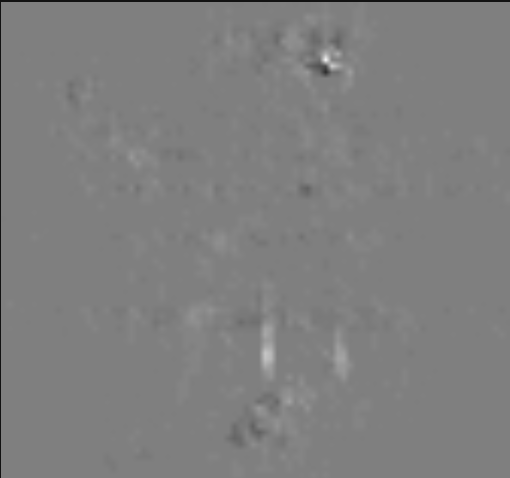}
        \includegraphics[width=0.19\linewidth]{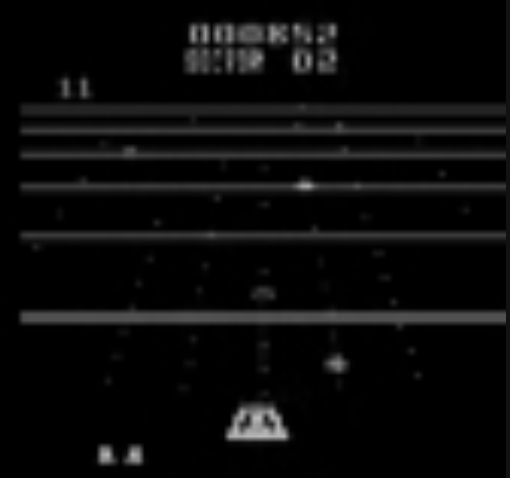}
        & \includegraphics[width=0.19\linewidth]{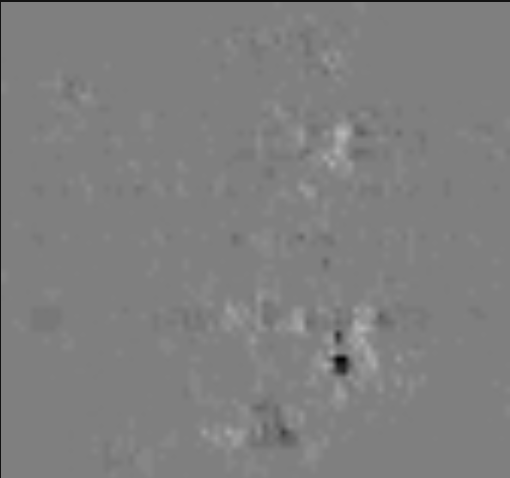}
        \includegraphics[width=0.19\linewidth]{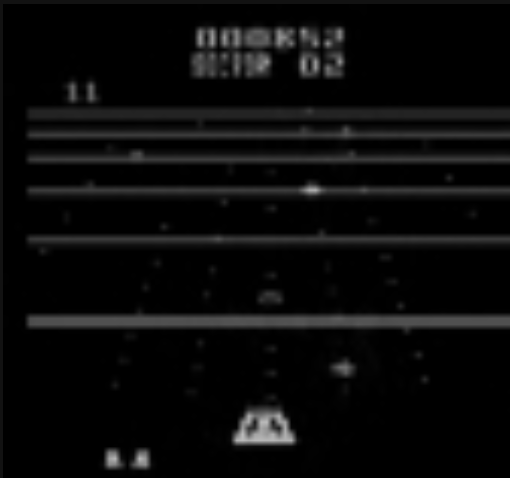}
        \\ \multicolumn{2}{c}{Adversarial explanations for ``Left'' and ``Right'' actions, original \gls{ae} evaluation.}
        
        \\ \includegraphics[width=0.19\linewidth]{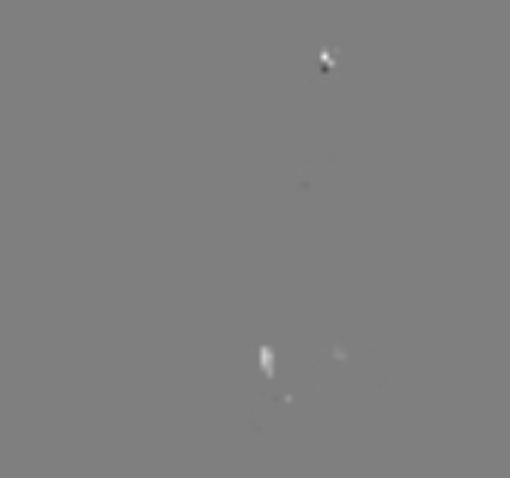}
        \includegraphics[width=0.19\linewidth]{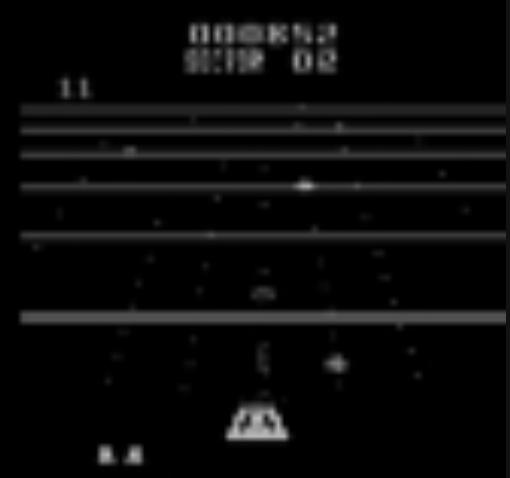}
        & \includegraphics[width=0.19\linewidth]{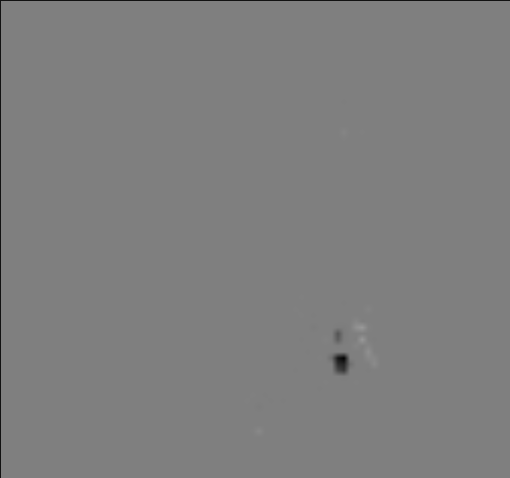}
        \includegraphics[width=0.19\linewidth]{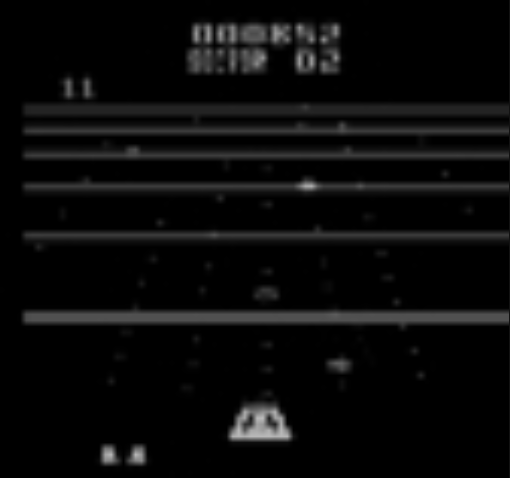}
        \\ \multicolumn{2}{c}{Adversarial explanations for ``Left'' and ``Right'' actions, sparsity-promoting evaluation (\cref{sec:app:ae:sparsity}).}
    \end{tabular}

    \vspace*{0.5em}
    \caption{Case study of analyzing a decision with an obstacle not aligned with gridlines. Here, the \gls{ssa} reconstructions clearly indicate non-reliance on specific gridline positions.}
    \label{fig:results:s4}
\end{figure}

This scenario is demonstrated in \cref{fig:results:s4}.

\subsection{Model Robustness}

As with traditional \glspl{ae}, the \gls{rl}-enabled \glspl{ae} presented in this work provide protection against adversarial input manipulations. To get reasonable \gls{rl} agent performance, we used lower regularization strength than in the original \gls{ae} paper. While we have not yet measured quantities comparable to the original work's \gls{ara} metric \cite{woods2019adversarial}, we did measure the \gls{rmse} of the perturbation (the difference between the perturbed input and the original input) to achieve a given change in policy for both a plain network and one treated with \gls{ae} methodology. In our experiments, the modified network had roughly identical performance to the original network ($9135$ vs $9274$ average episode performance, respectively), while the required \gls{rmse} was $1.81\times$ higher for the modified network. This shows that the modified network should be significantly more resistant to adversarial tampering, also implying that the \gls{ara} would be significantly higher, an exercise we leave for future work.

\section{Discussion and Future Work}

The provided capabilities should be a significant aid for humans trying to debug the \gls{ai} system's decisions or understand them in the context of a human/machine teaming scenario. We note that this work covered qualitative, case-study results only; specific human trials for quantitative results are left for future work.

As with classification-based \gls{ae}, the resulting explanations could also be passed to human subject matter experts for labeling, which in turn could improve the \gls{ai} model's robustness and decision making capabilities \cite{woods2019adversarial}. This is a promising future direction for this work, which would allow human feedback to improve the \gls{ai}'s decisions and explanations.

\section{Related Work}

Outside of \gls{rl}, a comparison of \gls{ae} to other state-of-the-art explanation methods has been presented in previous work \cite{woods2019adversarial}. For context within the field of \gls{xrl}, we refer to a few more recent studies. For example, Milani {\em et al.} \cite{milani2023explainable} suggested a taxonomy for \gls{xrl} that includes three scopes of explanation: Feature Importance, Policy-Level, and Learning Process. Within this framework, \glspl{ae} and \glspl{ssa} provide Feature Importance-level explanations, while safety margins and criticality provide Policy-Level explanations. 

One limitation of the \gls{rl} methods discussed in this work is that any given trained model only performs tasks in a single environment. That is, the methods do not provide foundation models in the sense that has been popularized by ChatGPT \cite{chatgpt}. However, there are efforts to build foundation-style models for \gls{rl} agents \cite{reed2022generalist}. We note that the methodologies proposed in our work are deliberately general; that is, we imagine that when methods for successful foundational \gls{rl} agents are developed, the techniques provided would also help with integrating them into combined human/\gls{ai} decision workflows and subsequent agent improvement.

\section{Conclusion}

As both the quantity of data available to decision makers and the complexity of decision environments increases, tools are needed to help humans meaningfully understand those environments and make effective decisions. Traditional decision assistance tools -- \glspl{dst} -- have focused on using data to help users answer questions they already know to ask. Instead, we proposed \glspl{idt} that use \gls{ai} methods to help users find the answer to a much broader set of questions, relying less on user knowledge and giving a broader perspective on the available data. These capabilities were provided through: \gls{rl} extensions for a MuZero-style agent for better learning and to reveal information about the environment to users; the extension of \gls{ae} to \gls{rl} to help users better understand different interpretations of available sensor data that would lead to different decisions; \glspl{ssa} to help users identify which data the \gls{ai} considers critical to understanding the current strategic scenario; and criticality and safety margins to help users understand the long-term consequences of different decisions. By combining all of these innovations, \glspl{idt} become a viable approach for joint human/machine decision making, and in particular, help to uncover unknown known/unknown factors. By focusing on not just \gls{ai} methods that autonomously make good decisions, but also in using those methods to expose available information to human decision makers, a platform is provided for making better decisions for critical missions.

\acknowledgments 
 
This material is based upon work supported by the Air Force Research Laboratory (AFRL) under Contract No. FA8750-22-C-1002.  This paper was PA approved under case number AFRL-2024-1725.

\section*{SBIR DATA RIGHTS}

Contract No.: FA8750-22-C-1002\\
Contractor Name: Galois, Inc.\\
Address: 421 SW Sixth Ave., Suite 300, Portland, OR 97204\\
Expiration of SBIR Data Rights Period: 01/18/2042\\
The Government's rights to use, modify, reproduce, release, perform, display, or disclose technical data or computer software marked with this legend are restricted during the period shown as provided in paragraph (b)(4) of the Rights in Other Than Commercial Technical Data and Computer Software - Small Business Innovation Research (SBIR) Program clause contained in the above identified contract. No restrictions apply after the expiration date shown above. Any reproduction of technical data, computer software, or portions thereof marked with this legend must also reproduce the markings.

\bibliography{report} 
\bibliographystyle{spiebib} 

\newpage
\appendix

\section{RL Agent Differences from MuZero}
\label{sec:app:rl}

In this section are the modifications to MuZero that were implemented and explored as part of this work.

\subsection{Balancing Pre-LayerNorm Activations}
Layer normalization, or LayerNorm \cite{ba2016layer}, is a way of normalizing the activity of different neurons, to ensure that any resulting activations are well-conditioned for learning. LayerNorm has a number of benefits, including similar training and test time behavior. However, in some contexts, LayerNorm does not perform as well as batch normalization (or BatchNorm) \cite{ioffe2015batch}. One theory for this that we explored is that there is no function forcing each data channel to actually be used. That is, LayerNorm enforces statistics over all data channels, whereas BatchNorm enforces statistics over each data channel independently. Other work has explored forcing the pre-LayerNorm activations of a \gls{nn} to ensure that each channel gets used via Wasserstein normalization \cite{joo2019regularizing}. However, we found that constraint to be too restrictive and difficult to balance; other optimization objectives were negatively impacted by the condition.

Instead, we take the idea of using gradients to condition the pre-LayerNorm activations \cite{joo2019regularizing} and use that idea to implement a variant of BatchNorm. To implement this, the forward pass is identical to LayerNorm, but during the backward pass, gradients pre-LayerNorm are modified. This is accomplished by swapping the channels (features) dimension with the batch dimension (assuming batch, channel, height, width ordering for images) and then computing statistics across the newly located batch dimension and all subsequent dimensions. Notably, at this step, we compute $\delta_{sd}$ as the standard deviation of each channel, and then multiply all of these by a scalar such that the new variance across all channels would be $1$. We found this $\delta_{sd}$ normalization to be important, as otherwise, the use of this method in, e.g., transformers, resulted in saturation of the attention layers. Finally, the rank of each value is computed within its statistics group (along the batch and spatial dimensions), and these ranks are converted to locations on a Gaussian via the inverse error function:

$$
    target = \delta_{sd}\sqrt{2}erf^{-1}\left[\frac{2R}{1 + N} - 1\right],
$$

\noindent where $R$ is the rank and $N$ is the number of samples in the statistics group. Finally, the gradient is modified such that all values are pulled toward their corresponding $target$ by a constant (we used $1e-2$) times the distance to $target$.

The result, loosely named LayerNormRebalanced, ensures that each feature channel is fully used across different inputs, while also retaining train/test computational similarity and not saturating attention layers affected by the change in gradient.

\subsection{Gaussian Mixture Models for Distributional RL}
\label{sec:meth:gmm}
MuZero only tracked expectations of rewards. For environments with multiple distinct reward states (e.g., a player ship destroying a target, merely surviving, or itself being destroyed), it can be helpful for a human co-decider to understand the risks of these different reward states. To capture this information, we turn to distributional \gls{rl} methods.

Briefly, distributional \gls{rl} involves capturing the entire reward distribution, instead of just an expectation. The original implementation that the authors are aware of was by Bellemare \etal, and used point masses to approximate the distribution over a fixed range \cite{bellemare2017distributional}. That technique has since been expanded with Implicit Quantile Networks \cite{dabney2018implicit} and \gls{gmm} approaches \cite{choi2019distributional,nam2021gmac}.

We appreciated the computational simplicity and representational power of the \gls{gmm} approaches, but found difficulties with the optimization of those representations \cite{choi2019distributional,nam2021gmac}. This tended to manifest in environments with large reward values -- e.g., a reasonably skilled Beamrider agent might have an expected total discounted reward range of $[0, 800]$, whereas a Pong agent might only have a range of $[-2, 2]$. Instead, we found success with a novel, rank-based \gls{gmm} update rule. Briefly, our representation and update rules are as follows.

For each distribution to be captured, decide a fixed number $N_{gaussian}$ of Gaussians to track, and output $3N_{gaussian}$ parameters from the corresponding value/reward network for the mean, standard deviation, and log weight of each Gaussian component. To update the distribution with a new sampled value, rank the output Gaussian means according to absolute distance from the new value. Calculate update weight factors for each individual Gaussian as $w_{rank=1} = 1$, $w_{rank\neq 1} = 1e-3$. 

To update the means of all Gaussians, pull them all toward the new value using \gls{mse} weighted by the weight factor multiplied (via a cross product) with the weight of the Gaussians themselves.

To update the standard deviations of all Gaussians, use \gls{mse} to pull them all toward the newly predicted variance plus the variance between the predicted mean and the new value. As with the mean, weight this pull by the weight factor crossed with the weight of the Gaussians themselves, but divide by 10 as standard deviation provides utility only for the user, and not for the algorithm's ability to make good decisions. Note that this does {\em not} produce an exact estimate of standard deviation -- however, what it does do is keep the {\em scale} of the gradient update corresponding to standard deviation on the same scale as the mean update. We found this to be the main issue with, e.g., GMAC \cite{nam2021gmac}. Essentially, its updates were not well balanced with all of the other updates required for a complex system like MuZero to work well.

To update the log weight of each Gaussian, we also take steps to balance the scale of this update against the scale of the other updates. Before computing the log softmax of all Gaussian weights, we pass the raw values output from the \gls{nn} through a hook that, on the backwards pass, weights each gradient by the absolute value of the difference between the overall expected mean value of the distribution and the expected value of the Gaussian being updated by that gradient. Since log softmax gradient updates range strictly on $[-1, 1]$, this multiplication brings them into the same scale as the mean and standard deviation updates. To prevent value drift, we also apply a mild centering effect such that the expected network output over the emitted Gaussian weight parameters is zero. In the forward pass, after this hook, the log softmax update is computed according to normal categorical cross entropy, where the log of each Gaussian's weight is increased in proportion with its corresponding weight factor.

\begin{figure}[b]
    \centering
    \includegraphics[width=\linewidth]{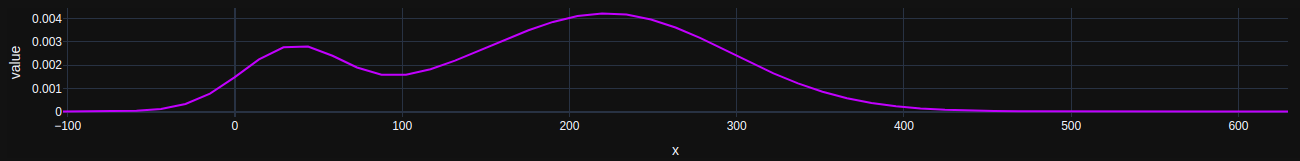}
    \caption{Screenshot of captured reward densities from Beamrider, in a situation where the user has a chance of losing the game (i.e., of zero reward).}
    \label{fig:meth:gmm}
\end{figure}

Put together, the magnitude of each component's update gradient is proportional to the expected L2 difference between the closest Gaussian mean and the sampled data point. An example captured distribution is shown in \cref{fig:meth:gmm}.

\subsection{Loss Segment Scaling}

For maximum diversity in different environments, and to increase interpretability for the user, we opted to keep reward units in the environment's scale (e.g., the number of points in a game), without clipping or other transformations. In a similar spirit as the gradient corrections from \cref{sec:meth:gmm}, we found that environments with large reward ranges needed additional loss tweaking from the standard MuZero format to support e.g. the SimSiam loss from EfficientZero \cite{ye2021mastering}. This is a direct result of the gradients for the value/reward/action networks scaling with the observed reward magnitudes, while other gradients do not scale with those quantities. The best solution we found for this was to estimate the L1 or L2 distance between actions chosen uniformly at random. This essentially is a measure of how sensitive the reward space was to different action choices, a quantity that roughly scales with the loss magnitudes used by the core MuZero mechanics; by dividing any reward-space updates (such as value/reward network updates) by this quantity, we found that the other network hyperparameters generalized across multiple environments.

\subsection{Complex Action Spaces via a Pairwise Policy Update}
\label{sec:meth:policy}

The original MuZero paper provided policy update and policy search methods that could only handle discrete action spaces. While an extension was developed for complex action spaces \cite{hubert2021learning}, we felt that that approach had a number of potential deficiencies. First, their inner policy update still uses visit counts. What happens when very similar actions (due to, e.g., a continuous action space) are visited twice as a result of the outer random sampling? Then, each has its own visit count from the \gls{mcts} process, despite being virtually identical, biasing toward over-sampling (and thus preferring) those values. This can be corrected via techniques such as importance sampling \cite{kloek1978}, but we wondered if there is a more straightforward way to produce policy updates.

Drawing inspiration from prior work on \gls{rl} for grammatical inference (RL-GRIT) \cite{woods2021rl}, we leveraged a pairwise policy update approach. This has a number of potential advantages: (1) as with \cref{sec:meth:gmm}, loss scales according to changes in expected reward, (2) it works the same for any kind of action space, and (3) as discussed in \cref{sec:meth:particles}, it lends itself very well to an alternative of \gls{mcts} that is faster to compute while conferring similar benefits.

Generally, the pairwise policy update works as follows: sample $N_{update}$ potential actions from the current action distribution, and for each $i \in 1..N_{update}$ compute both $\log P_{i,update}$, the log probability of selecting that action, and $Q_{i,update}$, the expected Q value for taking that action. Then, compute $\partial \log P_{i,update} = \frac{1}{N_{update}} \sum_{j=1}^{N_{update}} Q_{j,update} - Q_{i,update}$.

This formulation of the pairwise policy update has a few interesting properties. Gradients added to each log probability scale with $P(1 - P) |\Delta Q|$, meaning both that updates are rare for saturated actions, and that the scale of these updates is proportional to the expected difference in reward for making such a policy update. Additionally, because this is a symmetric loss, two actions with stochastic $Q_{update}$ values that have the same expectation will stay at exactly the same relative probability to one another. Since we expect the reward and value networks of a MuZero-style agent to be noisy (via optimization, not explicitly as in RainbowDQN \cite{hessel2018rainbow}), we find this property to be advantageous. 

The downside to this update rule is that it gives probability mass to actions that already have probability mass, and thus does not do a great job on its own with the exploration/exploitation trade-off.  We address this below.

\subsection{Directly Controlling the Exploration/Exploitation Trade-Off via a Pairwise Policy Update}
\label{sec:meth:explore}
The exploration/exploitation trade-off is a classic concern with \gls{rl} agents \cite{fan2022generalized}. Essentially, the agent does not begin its learning process understanding the world, and thus a certain amount of exploration is required. However, as the agent learns those world dynamics, it must shift to exploitation to achieve a high score. Often, in doing so, it reaches increasingly complex world states, which again necessitates some level of exploration until those new states are more fully learned. This teeter-totter is difficult to balance, with a number of successful approaches like Random Network Distillation \cite{burda2018exploration} and the more recent Generalized Data Distribution Iteration (GDI) \cite{fan2022generalized} tackling this through novelty rewards and/or bandit optimization.

Due to our pairwise policy update, we explored methods of handling this trade-off while also preventing excessive saturation of log probabilities. The final approach works as follows: $N_{update,uniform}$ actions are sampled uniformly at random from the entire action space, and corresponding $\log P_{i,update,uniform}$ values are computed. Then, the probability mass is given from the policy samples to these uniform samples. Importantly, the magnitude of this transfer is controlled by an integrating controller such that the frequency of policy samples associated with a sampling probability that is less than that of a uniform sampling probability is equal to $\rho_{target} = 0.1$. For easily testing different $\rho_{target}$ values, we scale the speed of the integrator update by $0.5 / \rho_{target}$.

Intuitively, this pushes states with little difference in $Q$ values between actions toward a uniform distribution, while allowing those states with significant $Q$ value differences to saturate. We found that this setup is very flexible, with the same hyperparameters working well in multiple environments. While the exploration/exploitation trade-off would likely perform better by using additional techniques like GDI \cite{fan2022generalized}, we found its performance in isolation to be satisfactory for many use cases.

\subsection{Particle Swarm Tree Search}
\label{sec:meth:particles}

\Gls{mcts}, used by MuZero, works well but depends heavily on tracking visit counts. In complex action spaces, this presents issues, as noted in \cref{sec:meth:policy}. Furthermore, \gls{mcts} is necessarily iterative in nature, and cannot benefit from the full parallelization offered by modern \glspl{gpu}.

To work around this, we devised a novel approach inspired by particle filter \cite{carpenter1999improved} and particle swarm optimization \cite{kennedy1995particle} approaches. 

To begin, $N_{ps}$ particles are sampled, with each particle following the agent's learned policy. Each of these particles are simulated for $T_{ps}$ time steps, at each step again following the learned policy from the newly computed world state according to the dynamics function. Observed immediate rewards and future Q values are accumulated as in MuZero's implementation of \gls{mcts}. This yields a set of particles which have both an action sequence and an expected Q value for following that action sequence from the current point in time.

Having collected $N_{ps}$ particles, we have $N_{ps}$ potential action trajectories and Q values. Like the original \gls{mcts} implementation, we need a means of determining an action distribution that combines both the estimated Q value and the likelihood of each action into an upper confidence bound. Unfortunately, in a similar issue to that mentioned in \cref{sec:meth:policy}, we do not have access to the frequencies of combined samples. That is, while we can use the log probabilities of each action sample, those samples will also come from that same distribution, meaning that direct use of the probability information has a sort of doubling effect. Instead, we can use a simple trick: if we add uniform noise, scaled by $c_1(\max_{ps}{Q} - \min_{ps}{Q})$ over all particles, to each particle's estimated Q value, and then take the top $\hat{N}_{ps} << N_{ps}$ of particles sorted by the modified Q values, then we have a distribution captured from the confidence bound. Here, $c_1 = 0.2$ is a constant determining the size of this effect. By adding uniform random noise, action samples with a larger sample probability are also more likely to receive a larger random sample. Thus, the probability of each sample is implicitly rolled into the newly sampled distribution, creating an altered action distribution in the same style as the \gls{mcts} approach.

This approach is certainly faster than \gls{mcts} (with a speedup factor of approximately $10$), and in our limited trials, equally effective.

\subsection{Modified MuZero Value Representation}

Our final modification to the MuZero family of \gls{rl} agents is to always have the Q network's result {\em include} the reward network's result for a single dynamics step. While our interpretation of MuZero is that it always has the Q network predict future rewards {\em after} the dynamics step, we found this much more difficult to train. The most likely explanation for this is that the dynamics function begins as nonsense, and when learning the initial \gls{wm}, it is easiest to learn a Q function that does not fully rely on an accurate dynamics model.

\section{Additions to Adversarial Explanations}
\label{sec:app:ae}

As mentioned in the main text, several changes to \gls{ae} were required to achieve satisfactory performance with \gls{rl}. These are described below.

\subsection{Gradient Scale Correction}

Gradient scale correction required two considerations: first, the massive difference in scale for outputs associated with rewards and action probabilities, and second, considerations for the magnitude of the second derivative required for \gls{ae} conditioning.

To handle the difference in output scales, we note that the input to any layer (including the final layer) has roughly standard normal statistics, as a result of the LayerNorms in the network. For large magnitude outputs, such as reward, this means that the final layer's weights are rather large. However, as mentioned earlier, this means that any gradients propagated backwards from this final layer would also be magnified, conflating the importance of different factors within the network. To fix this, we divide the gradient propagated backward from these final output layers by the expected standard deviation from the layer's weight matrix (the square root of the sum of squares). This expected standard deviation can be grouped over multiple output channels where they contribute to outputs in the same dimension; for example, with the \glspl{gmm}, the means, standard deviations, and weights constitute three dimensions which are spread over a greater number of parameters.

To handle the introduction of second derivatives into the loss function, it was crucial that this quantity be well-balanced amongst the different magnitudes of output from, e.g., the policy, Q, and reward networks from the MuZero agent. We found that anywhere a gradient was scaled within the network (e.g., during the dynamics update for MuZero or in the aforementioned output gradient scale correction), it was best to apply that gradient scaling only through a detached version of the original gradient. That is, for a gradient $g$, desired scale $s$, and assuming $d(g)$ means a detached version of the gradient $g$, we substituted the gradient $\hat{g} = g + d(g) * (s - 1)$. This detachment, coupled with pre-gradient scaling of the parameter by $1/s$ before the gradient operation and $s$ before adding it to the final loss function, was able to balance all of the magnitudes involved. By balance, we mean that the magnitude of the second gradient's effect gets scaled the same as the first gradient's effect as it passes through these different scalars and multiples. For example, as MuZero scales each dynamics update's gradients by $0.5$, all output values (before the gradient for \gls{ae} conditioning) were scaled by $2^\tau$, and after the gradient loss was computed, this factor was divided out again.

An additional correction was needed for the policy update, as the policy loss scales down as the action distribution becomes saturated. To account for this, the scale on smoothing for policy parameters was scaled by $0.25\rho_{target}$ from \cref{sec:meth:explore}. 

\subsection{Optimizer Scale Correction}

Our experiments exclusively use the Adam optimizer for its fast convergence properties \cite{kingma2014adam}. However, when used in combination with \gls{ae} for \gls{rl}, we noticed that learning was often unstable, with the occasional massive exploding gradient. To fix this, we used a variant of Adam that was step-size aware; that is, in the ``Update Parameters'' step, if $\left|\hat{m}_t / (\sqrt{\hat{v}_t} + \epsilon)\right| > (c_{maxstep} = 100)$ for any parameter update, then all parameter updates were scaled down so that the max step size was 100 (which is then multiplied by the learning rate). This solved such instabilities, and retained good optimizer quality.

\subsection{Controlling Regularization Strength}

The original \gls{ae} work controlled the \gls{ae} regularization loss strength by targeting a given training loss \cite{woods2019adversarial}. We did not find this to be effective for \gls{rl}, as training loss is unpredictable and varies significantly by environment. One approach would have been to target an average agent reward, but that is a complex function of both the environment and the agent regularization itself. Instead, we target an expected L1 magnitude of the gradient for the chosen action's log probability with respect to the inputs. This effectively bounds the Lipschitz constraint from \gls{ae} to enforce a given rate of change, which we found to be both easy to adjust and surprisingly consistent across environments.

\subsection{Sparsity-Promoting Gradient-Minimization Function}
\label{sec:app:ae:sparsity}
We used $z=1.25$ for the $L_{adv,z,q}$ function from Section III-B of the original \gls{ae} paper \cite{woods2019adversarial}. This encouraged sparser gradients, which was convenient for human interpretation in many of the environments we tried. During evaluation (not training), we also found it useful to adjust the manner in which \glspl{ae} were generated. By adding an additional L1 loss term back toward the original image, the explanations were even sparser, allowing for easier interpretation.

\section{Additional Figures}

\begin{figure}[t]
    \centering
    \includegraphics[width=\linewidth]{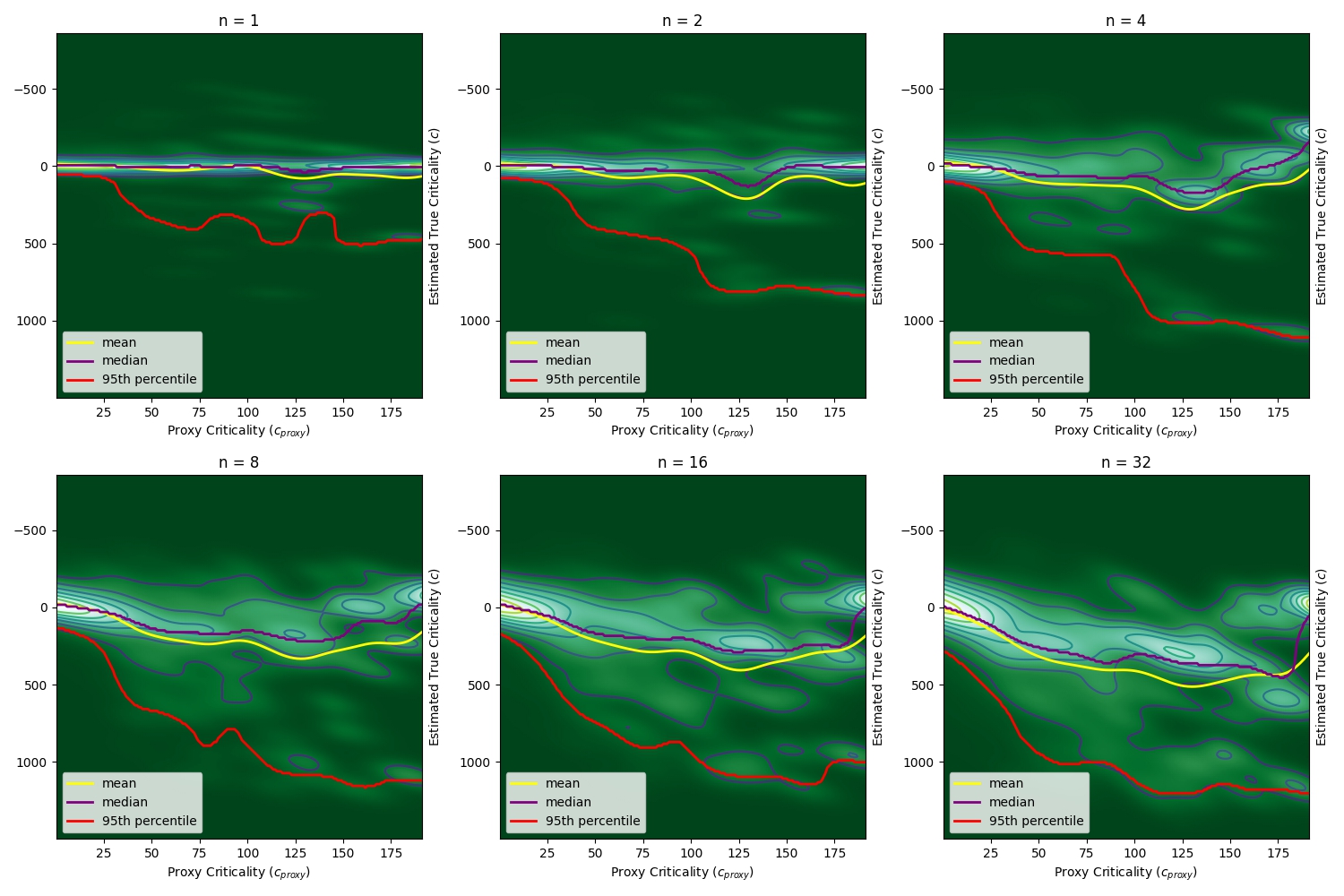}
    \caption{Kernel density plots that capture the relationship between true criticality (y-axis) and proxy criticality (x-axis), for different numbers $n$ of random actions. See Grushin {\em et al.} for additional information on interpreting this figure \cite{grushin2023safety,grushin2024criticality}.  In particular, note that the 95th percentile curves become the boundaries between the shaded regions in \cref{fig:results:model:sm}, though they are first adjusted such that they are monotonic with respect to proxy criticality.}
    \label{fig:app:model:kde}
\end{figure}

\end{document}